\documentclass{article}

\pdfoutput=1

\usepackage[final, nonatbib]{neurips_2024}

\usepackage[numbers]{natbib}
\usepackage[utf8]{inputenc} 
\usepackage[T1]{fontenc}    
\usepackage{url}            
\usepackage{booktabs}       
\usepackage{amsfonts}       
\usepackage{nicefrac}       
\usepackage{microtype}      
\usepackage{xcolor}         
\usepackage{lipsum}
\usepackage{fancyhdr}       
\usepackage{graphicx}       
\usepackage{multirow}
\usepackage{amsmath}
\usepackage{amssymb}
\graphicspath{{media/}}     
\usepackage{array}
\usepackage{subfigure}
\usepackage{placeins}
\usepackage[ruled, vlined]{algorithm2e}
\usepackage[title]{appendix}
\usepackage[pdftex, pdftitle={Tabular Transactions}]{hyperref}

\title{Scalable Representation Learning for \\ Multimodal Tabular Transactions  }

\author{%
  Natraj Raman, Sumitra Ganesh and Manuela Veloso \\
  JPMorgan AI Research\\
  \texttt{first.last@jpmorgan.com} \\
}

\begin{document}

\maketitle

\begin{abstract}
Large language models (LLMs) are primarily designed to understand unstructured text. When directly applied to structured formats such as tabular data, they may struggle to discern inherent relationships and overlook critical patterns. While tabular representation learning methods can address some of these limitations, existing efforts still face challenges with sparse high-cardinality fields, precise numerical reasoning, and column-heavy tables. Furthermore, leveraging these learned representations for downstream tasks through a language based interface is not apparent. In this paper, we present an innovative and scalable solution to these challenges. Concretely, our approach introduces a multi-tier partitioning mechanism that utilizes power-law dynamics to handle large vocabularies, an adaptive quantization mechanism to impose priors on numerical continuity, and a distinct treatment of core-columns and meta-information columns. To facilitate instruction tuning on LLMs, we propose a parameter efficient decoder that interleaves transaction and text modalities using a series of adapter layers, thereby exploiting rich cross-task knowledge. We validate the efficacy of our solution on a large-scale dataset of synthetic payments transactions.  
\end{abstract}
\section{Introduction}

Digital transactions facilitate the instantaneous exchange of payments across the globe, and analyzing them effectively requires sophisticated AI models. Transactions are typically organized as tabular data, comprising a mixture of categorical (e.g., currency), numerical (e.g., amount), and textual (e.g., account name) columns. Developing efficient representations for transaction records is crucial for a range of tasks~\cite{diadiushkin2019fraud}, including fraud detection, transaction tagging, and process optimization. 

Recent successes of Large Language Models (LLMs) have led to a growing temptation~\cite{hegselmann2023tabllm} to serialize tabular transaction data into a textual format to leverage the reasoning capabilities of these models. However, this approach comes with several disadvantages such as a loss of structural information, increased sequence length leading to computational inefficiency, difficulties with precise numerical reasoning and a lack of domain specific nuances. Instead, the potential of transformer architectures~\cite{vaswani2017attention} can be better realized by adapting them specifically to the distinctive characteristics of tabular data.

Several recent works~\cite{zhu2023xtab, ye2024ptarl} have proposed tabular representation learning by harnessing the strengths of transformers. However, these approaches often fail to address the unique challenges posed by transaction data. For example, transaction data includes millions of account identifiers. The conventional method of assigning a unique $D$ dimensional embedding vector to each categorical value will result in gigantic embedding tables. This not only increases the model size but also encounters issues with long-tailed distributions, where many identifiers are infrequent and lack sufficient data for effective parameter learning. Tokenizing these identifiers into sub-words is not beneficial, as unlike traditional English words, the subword components of identifiers do not possess meaningful semantic similarities. Transaction data also contains several auxiliary columns and including them directly during representation learning can lead to significantly increased sequence lengths. Furthermore, the standard tokenization of numerical values ignores their distributional continuity~\cite{thawani2021representing}. 

To address the aforementioned challenges, we propose a scalable approach for self-supervised transaction representation learning. Specifically, we introduce a partitioning embedder that leverages power-law distributions~\cite{newman2005power} to allocate the embedding space non-uniformly, granting larger subspaces to frequently occurring vocabulary items. Additionally, we distinguish between core columns and meta-columns, pre-learning the latter offline and employing summary representations to enhance modularity and computational efficiency. For numerical features, we implement adaptive quantization to improve resolution and mitigate noise. In order to encourage globally coherent representation space, we formulate a composite loss in a metric learning setting. 

Integrating tabular representations with a language model is crucial for efficiently executing downstream tasks. Traditional instruction tuning~\cite{zhang2023instruction} often involves billions of parameters, making it resource-intensive. Although parameter-efficient tuning methods~\cite{han2024parameter} reduce the number of parameters, they may fall short in sharing information across tasks, which is vital for effective knowledge transfer. To address this, we propose a scalable solution that keeps the parameters of both the tabular encoder and the LLM frozen. Instead, we introduce a set of adapter layers designed to efficiently handle variable number of rows, interleave the different modalities and learn cross-modal alignment. 

\begin{figure*}[tbp]
\centering
\includegraphics[width=\textwidth]{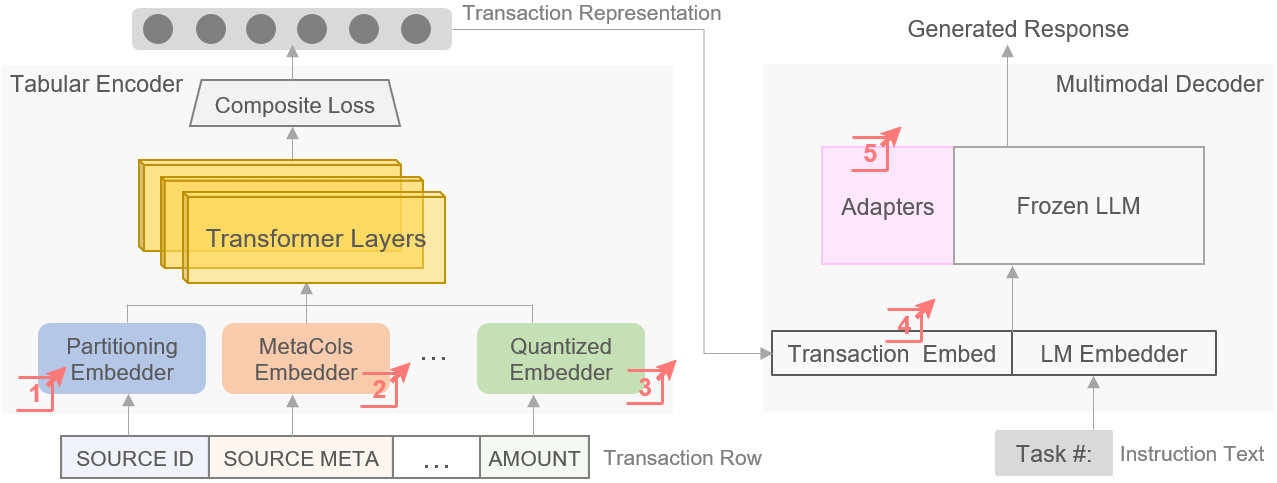}
\caption{Scalable multimodal foundation model for transaction data. (1) Sparse yet large vocabulary categorical features are learned using multi-tier partitions. (2) Meta-column features are pre-learned offline, with only selected segments integrated. (3) Numerical values are quantized into a coarse vocabulary to emphasize ranges.  (4) Compact transaction embeddings are interleaved with verbose instruction text. (5) A limited set of alignment parameters are fine-tuned for task-specific adaptation.} 
\label{fig_overview}
\end{figure*}

We conduct extensive experiments on a synthetic dataset comprising millions of transactions, characterized by complex account structures. In summary, our core contributions are: (i) a tabular encoder that scales seamlessly for large vocabularies, wide tables and granular numerical values. (ii) a multimodal decoder that utilizes a limited set of parameters to optimize downstream tasks. Figure ~\ref{fig_overview} provides an overview.

\section{Related Work}
Self-supervised representation learning for tabular data is an active research field~\cite{wang2024survey}, with recent works~\cite{gorishniy2021revisiting, wang2022transtab, yan2024making} focusing on pre-training paradigms common in transformer based architectures. Auto-encoder frameworks that reconstruct masked segments of tables is a popular approach~\cite{huang2020tabtransformer}, although auto-regressive approaches are not uncommon~\cite{zhang2023generative}. Some efforts~\cite{yin2020tabert, gardner2024large} also linearize tables into a text format. Despite attempts to pretrain on various types of tables~\cite{zhu2023xtab}, generalizing across tables remain a challenge. Embedding compression techniques to handle large sparse features have been explored ~\cite{kang2020learning, li2024embedding}, although they are more commonly used in recommender systems rather than tabular data. Tabular data has been integrated with other modalities~\cite{hager2023best}, but these approaches are tailored for specialized applications. In the absence of a universally applicable tabular foundation model~\cite{van2024tabular}, domain specific representation learning, such as ours, is often necessary. 
\section{Tabular Encoder}
Let $\mathbf{x}=(x_c)_{c=1}^{C}$ denote a tabular row with $C$ columns, and $\mathcal{D}=\{\mathbf{x}^i\}_{i=1}^N$ be the tabular dataset. The table may be semi-structured, containing  different types of columns, including a numerical column $x_c^{num} \in \mathbb{R}$, a categorical column $x_c^{cat} \in \mathbb{Z}^{V_{cat}}$ and a text column $x_c^{text}=(x_{c_1}, \ldots,x_{c_T})$, where a text is a sequence of tokens $x_{c_t} \in \mathbb{Z}^{V_{txt}}$. Here ${V_{cat}}$ and ${V_{txt}}$ denote the vocabularies of the categorical and text columns, respectively. For instance, in a transaction table, the transacted amount is an example of a numerical column, while the country and names of the transacting parties are examples of categorical and text columns, respectively. 

The primary objective of the encoder is to learn a function $f : \mathbf{x} \rightarrow \mathbb{R}^{C \times D}$ that maps the columns into a $D$ dimensional space. A compact representation for the entire record can be obtained by either pooling the representations across the columns or by maintaining a specific summary column. Most tabular representation learning methods treat the combined sequence of columns as a set of tokens and employ a language modeling approach ~\cite{devlin2018bert} to learn the representations. 

In this work, we address the challenges associated with columns that have large categorical vocabularies. Our approach also leverages factorizations inherent to tabular data, employs discretization techniques to incorporate domain knowledge, and avoids representation localization. These enhancements aim to improve the efficiency and effectiveness of tabular data representation learning.

\subsection{Partitioning Embedder}
In traditional language modeling, each token $V_i$ in the vocabulary of categorical items is assigned a unique embedding vector $e_i \in \mathbb{R}^D$ in the embedding space using an embedding matrix $E$ as follows: 
\begin{equation}
e_i = E[i,:] \quad \forall i \in {1, \ldots, V}, \quad E \in \mathbb{R}^{|V| \times D}.
\end{equation}

This approach presents a significant challenge for large vocabularies, as the number of parameters required for learning the embedding matrix increases linearly with the vocabulary size. 

\begin{figure*}[tbp]
    \centering
    \subfigure{
        \includegraphics[width=0.45\textwidth]{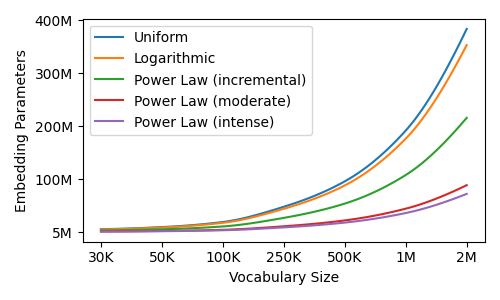}
    }
    \hspace{0.05\textwidth} 
    \subfigure{
        \includegraphics[width=0.45\textwidth]{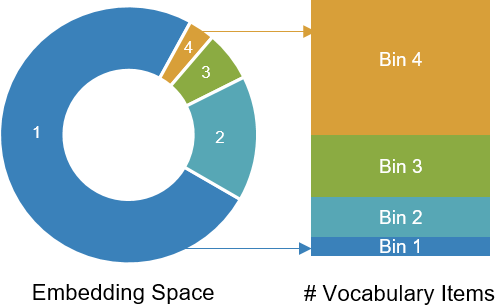}
    }
    \caption{Preferential partitioning of the embedding space. \emph{Left}: Embedding parameter size can increase dramatically with a large vocabulary. \emph{Right}: A non-uniform partitioning mechanism where a preferred bin with fewer vocabulary items is allocated a larger portion of the embedding space.     }
    \label{fig_partoverview}
\end{figure*}

We propose a scalable solution with reduced linear complexity. Specifically, we divide the vocabulary into $B$ bins such that $V = V^1 \cup \ldots \cup V^B$. A critical question arises: how should we assign the vocabulary items to the bins? The left side of Figure ~\ref{fig_partoverview} displays the number of embedding parameters required for different vocabulary sizes under various distributions for a fixed $D$ and $B$. We adopt a power-law strategy, as many real-world datasets exhibit power-law distribution in the frequency of vocabulary items. Consequently, the number of items in adjacent bins decreases rapidly, reflecting the steep drop-off characteristic of power-law distributions. Formally,
\begin{equation}
|V^b| = \frac{|V| \cdot b^{-\alpha_v}}{\sum_{j=1}^{B} j^{-\alpha_v}}  \quad \forall b \in 1,\ldots,B
\end{equation}
where the exponent parameter $\alpha$ controls the rate of decay. By assigning fewer items to bins corresponding to the most frequent vocabulary items, important items can receive more representation capacity and therefore achieve better scalability and robustness. 

Similarly, the embedding space is also divided into $B$ different subspaces\footnote{For simplicity, we assume the same number of partitions for both the vocabulary and embedding space.} using a power-law distribution, controlled by $\alpha_d$, such that $\mathbb{R}^D = \mathbb{R}^{D^1} \oplus \ldots \oplus \mathbb{R}^{D^B}$. Assuming $|V^1| \ll \ldots \ll |V^B|$ and $D^1 \gg \ldots \gg D^B$, the non-uniform partitioning mechanism assigns each item in $V^b$ a $D^b$ dimensional subspace. Thus, preferred bins are allocated a larger portion of the embedding space, as shown on the right side of Figure ~\ref{fig_partoverview}. In contrast to equation (1), we now have  
\begin{equation}
e_i^b = E^b[i,:] \quad \forall i \in {1, \ldots, V}, \quad E^b \in \mathbb{R}^{|V^b| \times D^b},
\end{equation}
and the number of embedding parameters reduces from $|V|D$ to $\sum_b{|V^b|D^b}$.

\subsection{Meta Column Representations}
Tabular data often contains meta-columns that provide additional contextual information about core columns. In practice, meta-columns, such as account-related attributes, are often stored in separate tables as part of database normalization. Instead of flattening these columns, it is preferable to learn their representations offline and utilize a summarized form in the tabular encoder. This approach offers several benefits: (a) An increase in the number of meta-columns does not affect the tabular encoder, thereby making the table representation learning process more scalable. (b) The modular structure allows for enhanced capturing of  intricate relationships within the meta-columns. (c) A pooled summary representation of these meta-columns can be used, thereby reducing the sequence length of the encoder.  

Formally, let $\mathbf{x_g} \subset \mathbf{x}$ denote a row with a subset of columns from $\mathbf{x}$. We first use a separate function $\Xi_g : \mathbf{x_g} \rightarrow \mathbb{R}^{C_g \times D}$ to learn the representations for this subset, and then augment their embeddings with those from other columns as 
\begin{equation}
embedding(\mathbf{x}) = (\Xi(x_1),\ldots, \Xi_g(\mathbf{x_g}), \ldots ,\Xi(x_C)),
\end{equation}
where $\Xi$ is an embedding function\footnote{We have slightly abused the notation here for brevity. $\Xi$ can be different for each column.}. By ensuring that $C_g < |\mathbf{x_g}|$, the sequence length used for table representation learning can be reduced.

\subsection{Numerical Quantization}
Language models use general tokens to represent numerical values, often resulting in subpar performance in tasks that require numerical reasoning~\cite{thawani2021representing}. While directly incorporating a continuous value can capture precise numerical information, they are more sensitive to noise and outliers and are challenging to scale. A quantized approach, which maps continuous values to discrete bins (e.g. 1028 and 1036 to 1000), and employs a custom vocabulary of numerical tokens offer a convenient alternative. It reduces the complexity of the input space, enhances robustness to noise and leads to more stable predictions. Formally, we define a numerical vocabulary $Q$ that adapts to resolution, featuring finer spacing for smaller numbers and progressively larger spacing for larger numbers. We assign a continuous value $y$ to token $Q_i$ as $\arg\min_{i} |x-Q_i|$. This adaptive quantization ensures the model captures essential numerical information while maintaining stability and scalability.

\subsection{Composite Loss}
The reconstruction loss employed in self-supervised learning excels at capturing the local structure of the data. However, it falls short in sharing information across samples, thereby neglecting the global structure. To overcome this limitation, we include an additional batch hard triplet loss~\cite{zeng2020hierarchical} term that considers the relative distances between samples within a batch. Specifically, the distance between similar samples are minimized while the distance between dissimilar ones are maximized. This composite loss formulation encourages our model to learn a globally coherent representation space, effectively capturing relationships across different rows in the table. In the scenario where explicit labels for similarity are unavailable, we generate two different perturbed versions of $\mathbf{x}$ to serve as similar examples. This strategy ensures that the model not only understands the fine-grained details of individual samples but also grasps the broader context, leading to a more robust and comprehensive understanding of the data.

\section{Multimodal Decoder}
Let $\mathcal{S}=\{(\{\mathbf{x}\}_{i,j=1}^{M_i}, \mathbf{t}_i, k_i, \mathbf{y}_i)\}_{i=1}^N$ be an instruction tuning dataset, where an example $i$ comprises a tuple of $M_i$ different transaction records $\mathbf{x}$, an instruction text $\mathbf{t}$, a task identifier $k \in {1...K}$ and a desired text response $\mathbf{y}$. Given a tabular encoder $f$ and a language model $\mathrm{LLM}$ with an embedding layer $\Xi_{\mathrm{LLM}} : \mathbb{Z}^+ \rightarrow \mathbb{R}^D$, the objective is to perform instruction tuning using $\mathcal{S}$. 

This setting presents several challenges compared to traditional instruction tuning: (a) An example might contain more than one transaction record. (b) The language model must learn to explicitly reference specific records when generating responses. (c) The parameters of the language model or the tabular encoder cannot be updated to ensure scalability and to prevent catastrophic forgetting. (d) Transaction records are not native to the language model, necessitating the learning of their alignment with the language model vocabulary. (e) It is essential to leverage commonalities across tasks. 

\begin{figure*}[tbp]
\centering
\includegraphics[width=\textwidth]{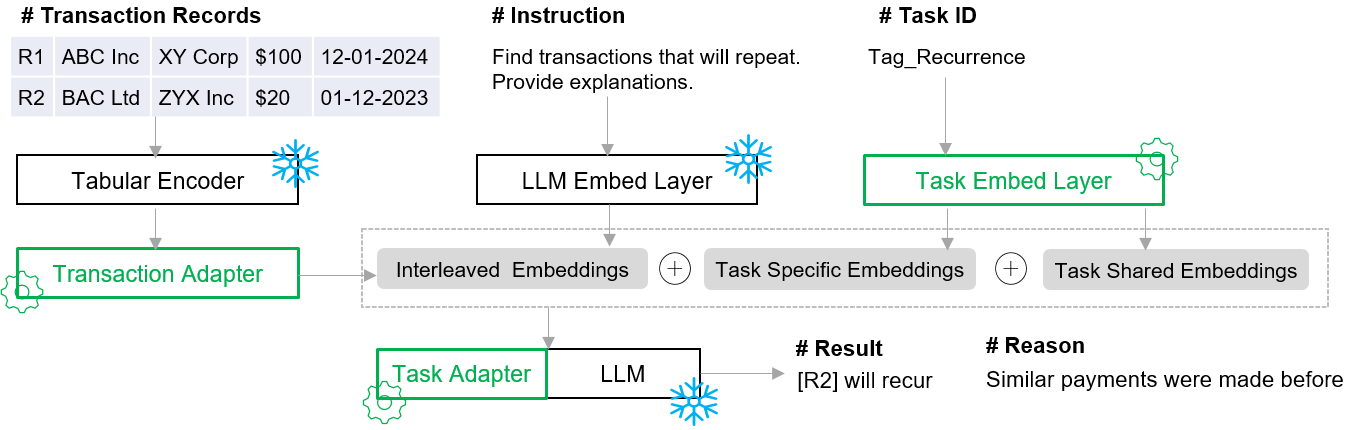}
\caption{Instruction tuning workflow. Transaction records, textual instructions, and task information are combined and fed into an LLM. All layers are frozen (shown in blue) except for the adapter layers (shown in green). These adapter layers learn to align the transaction and text embeddings with the specified task, enabling the LLM to generate response instructions. } 
\label{fig_decoder}
\end{figure*}

To address the first two challenges, we augment each transaction record with a row sentinel $s$ (e.g. $s(1)=[R1]$) that uniquely identifies a record. This sentinel is part of the language model vocabulary. Consequently, the inputs passed to the language model are interleaved with text tokens followed by transaction tokens, and then text tokens again and so on (see equation \ref{eq:decode}). The third challenge is tackled by freezing the parameters of both $f$ and $LLM$. The subsequent challenges are addressed by introducing new layers, which we detail below. Figure ~\ref{fig_decoder} provides an overview of the instruction tuning components.  

\subsection{Adapter Layers}
Instead of directly using the representation obtained from $f(\mathbf{x})$, we introduce a function $\phi$ to transform the tabular encoder representation. Typically, $\phi$ consists of a small set of transformer layers, and its parameters $\Phi$ are learned during instruction tuning. Intuitively, these transaction adapter layers aim to modify the transaction representations in a manner that the language model can comprehend, without altering the core structure captured by the original representations.

We also introduce a new embedding layer $\Xi_{task}: {1...K} \rightarrow \mathbb{R}^D$ with parameters $\psi$ to convert a task id $k_i$ into a task embedding. The task embedding contains subspaces unique to each task, as well as a subspace shared across all the tasks. Finally, we augment each layer of $\mathrm{LLM}$ with additional parameters $\varphi$, similar to prompt tuning~\cite{li2021prefix}.The following loss is minimized during instruction tuning:
\begin{flalign}
\label{eq:decode}
& \mathbf{z_i} = \Xi_{\mathrm{LLM}}(s(1)) \oplus \phi(f(\mathbf{x}_{i1})) \oplus \ldots \Xi_{\mathrm{LLM}}(s(M_i)) \oplus \phi(f(\mathbf{x}_{iM_i})) \oplus \Xi_{\mathrm{LLM}}(\mathbf{t}_i) \oplus \Xi_{task}(k_i) & \\
& \mathcal{L}_{\mathrm{LLM}} = - \sum_{i}{\log P(\mathbf{y}_i | \mathbf{z}_i; \Phi, \psi, \varphi )} &
\end{flalign}

\section{Experiments}

\begin{figure*}[tbp]
\centering
\includegraphics[width=\textwidth]{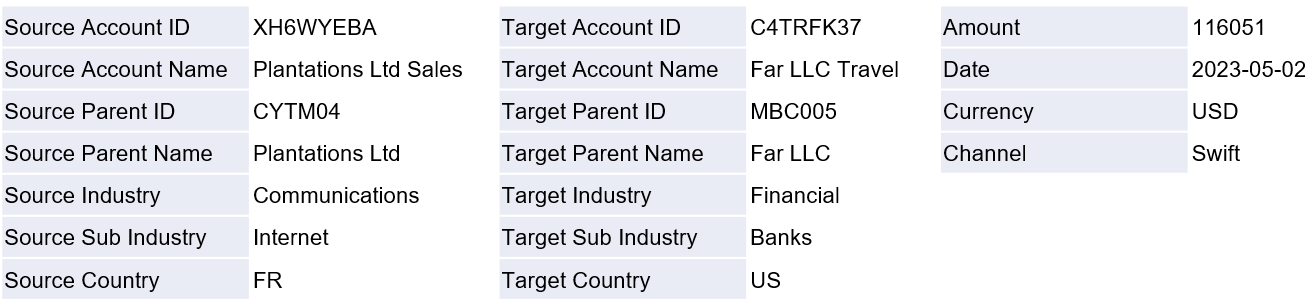}
\caption{Sample synthetic transaction record. } 
\label{fig_synthdata}
\end{figure*}

\begin{table*}[t]
\addtolength{\tabcolsep}{-1pt}
\caption{Pretraining results for tabular encoder showing reconstruction accuracy for different model variants. The source and target fields of a transaction are combined for brevity. }
\centering
\hspace*{-0.5cm}
\begin{tabular}{l|l|rr|rrr|r}
\addlinespace[0.5ex]
\specialrule{0.8pt}{0pt}{0pt}
  & & \multicolumn{2}{c|}{\small\textbf{Large Vocab}} & \multicolumn{3}{c|}{\small\textbf{Categorical}} & \multicolumn{1}{c}{\small\textbf{Numeracy}} \\
  \rule{0pt}{2.5ex} \small\textbf{Model Type} & \small\textbf{Size} & \small\textbf{Account ID}  & \small\textbf{Parent ID} & \small\textbf{Industry} & \small\textbf{Country} & \small\textbf{Currency} & \small\textbf{Amount}\rule[-1.2ex]{0pt}{0pt}  \\
\specialrule{0.8pt}{0pt}{0pt}
\addlinespace[0.5ex]
Partitioned Embeds & 100M & 54.5 & 57.7 & 83.1 & 67.4 & 83.8 & 55.3 \\
Classical Embeds & 185M & \underline{58.7} & \underline{61.8} & \textbf{84.9} & 70.4 & \textbf{86.9} & \textbf{60.4} \\
Partitioned + MetaCols & 100M & \textbf{60.7} & \textbf{62.6} & \underline{84.7} & \textbf{76.3} & \underline{85.8} & \underline{60.2} \\
Classical Loss & 100M & 46.3 & 51.6 & 81.2 & \underline{70.5} & 80.4 & 48.1 \\
Autoregressive Mask & 120M & 53.1 & 48.6 & 80.6 & 67.3 & 80.7 & 49.8 \\
\addlinespace[0.5ex]
\specialrule{0.8pt}{0pt}{0pt}
\end{tabular}
\label{tab_pretrainres}
\end{table*}

\subsection {Transactions Dataset}
Our experiments focus on large-scale financial transactions conducted between businesses. These transactions are characterized by high value, substantial volume, and complex account structures, necessitating a precise and nuanced understanding. Due to the sensitive nature of real data and the extensive volume of information involved, we utilize a synthetically constructed dataset comprising 10 million transaction records. A sample record is illustrated in Figure ~\ref{fig_synthdata}. Of particular interest are the Account ID and Parent ID fields, which together encompass a vocabulary of nearly 125,000 items.

\begin{algorithm}
\caption{Generation process for the synthetic transactions}\label{alg:synthdata}
\SetKwComment{Comment}{$\triangleright$\ }{}
\SetKwInOut{Require}{Require}
\Require{Number of Parent Companies $C$, Number of Transactions $T$}
\Require{Models $\mathrm{M_{Comp}, M_{Dest}, M_{Txns}, M_{Amount}, M_{Date}}$ to sample from}
\Require{Function $\mathrm{CreateAccount}$ that creates a synthetic account with name, country, etc.}
\SetKwFunction{CreateAccount}{CreateAccount}
\SetKwFunction{Sample}{Sample}
\SetKwFunction{Size}{size}
\SetKwFunction{Append}{append}

$accs \gets []$\;
\For{each $c$ in $C$}{
    $N_c \sim \mathrm{M_{Comp}}$ \Comment*[r]{Sample number of accounts in a company}
    \For{each $n$ in $N_c$}{
        $accs \gets accs \oplus \CreateAccount()$\;
    }
}
$txns \gets []$\;
\While{$\Size{txns} < T$}{
    $src \sim accs$ \Comment*[r]{Sample a source account}
    $N_d \sim \mathrm{M_{Dest}}$ \Comment*[r]{Sample number of target accounts}
    \For{each $n$ in $N_d$}{
        $dest \sim accs$ \Comment*[r]{Sample a target account}
        $N_t \sim \mathrm{M_{Txns}}$ \Comment*[r]{Sample number of txns between source and target}
        $\mu_a, \sigma^2_a \sim \mathrm{M_{Amount}}$ \Comment*[r]{Sample mean and variance of amount}
        $\mu_d, \sigma^2_d \sim \mathrm{M_{Date}}$ \Comment*[r]{Sample mean and variance of dates}
        \For{each $t$ in $N_t$}{
            $amt \sim \mathcal{N}(\mu_a, \sigma^2_a)$ \Comment*[r]{Sample amount}
            $dte \sim \mathcal{N}(\mu_d, \sigma^2_d)$ \Comment*[r]{Sample date}
            $txns \gets txns \oplus (src, dest, amt, dte)$\;
        }
    }
}
\end{algorithm}

\begin{figure*}[t]
\centering
\includegraphics[height=4cm]{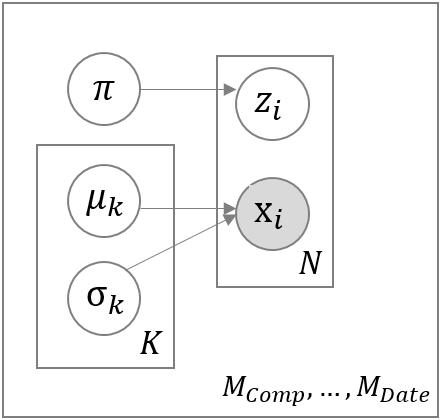}
\caption{Mixture models in plate notation. These models are used for sampling the number of accounts in a company, the number of accounts a given account can transact with, the number of transactions between a (source, target) pair, and the amount and date values.} 
\label{fig_synthgen}
\end{figure*}

The synthetic transaction generation process is outlined in Algorithm ~\ref{alg:synthdata}. The algorithm takes as input the number of parent companies for defining the account structure and the number of transactions to generate. It further requires a function that can create the account columns (e.g. account names, country etc.) and a set of probabilistic models to sample from.  For instance,  $\mathrm{M_{Comp}}$ models the number of accounts corresponding to a company using a Gaussian Mixture Model with $K$ components that can capture the variations in account structure across different types of companies. Similarly, $\mathrm{M_{Dest}}$  models  the number of target accounts a source account can transact with, $\mathrm{M_{Txns}}$ the number of transactions between a (source, target) pair, and $\mathrm{M_{Amount}}$ and  $\mathrm{M_{Date}}$ the amount and date values respectively. Figure ~\ref{fig_synthgen} describes the variables involved in plate notation.  

\begin{figure*}[t]
\centering
\includegraphics[width=\textwidth]{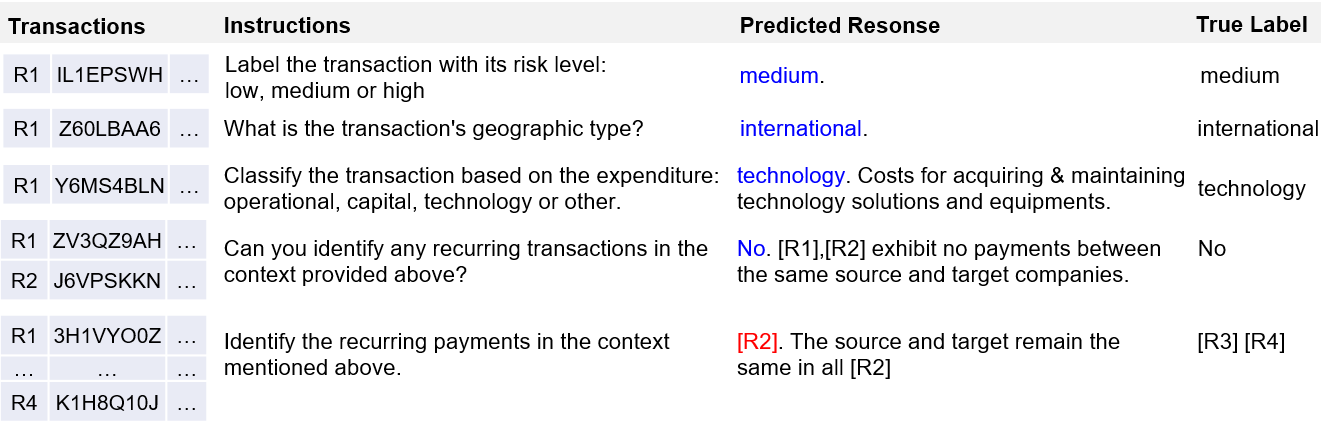}
\caption{Qualitative examples for risk, geo, expense and recurrence tagging tasks. Some tasks (e.g. recurrence) accept multiple transaction records and use row sentinels (e.g. [R2]) to identify them. Correct answers are highlighted in \textcolor{blue}{blue} while incorrect answers are shown in \textcolor{red}{red}.  } 
\label{fig_qual}
\end{figure*}

For instruction tuning, we generate 100,000 template-based instructions and corresponding desired responses for four distinct tasks. These tasks include tagging a transaction with its risk level (Low, Medium, or High), geographic span (US, Americas, EMEA, Asia, or International), expense type (Capital, Operational, Technology, or Other), and recurrence category (Yes/No and identifying specific transactions). Figure ~\ref{fig_qual} presents a few sample instructions and responses.

\begin{figure*}[t]
    \centering
    \subfigure{
        \includegraphics[width=0.45\textwidth]{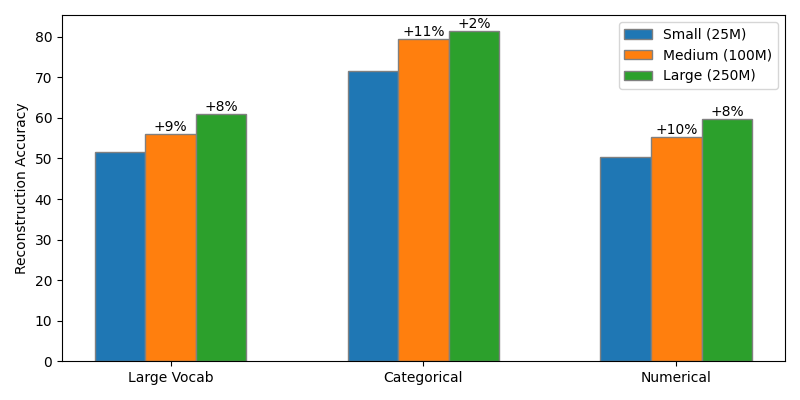}
    }
    \hspace{0.05\textwidth} 
    \subfigure{
        \includegraphics[width=0.45\textwidth]{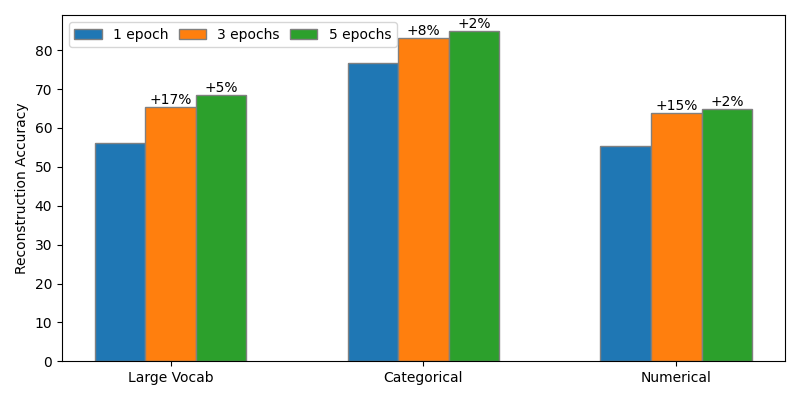}
    }
    \caption{Ablation study of reconstruction performance across different model sizes and training durations. The relative gain over the previous counterpart is annotated. \emph{Left}: Small, medium and large model sizes. \emph{Right}: 1, 3 and 5 training epochs. }
    \label{fig_epochs}
\end{figure*}

\begin{figure*}[t]
\centering
\includegraphics[height=5cm]{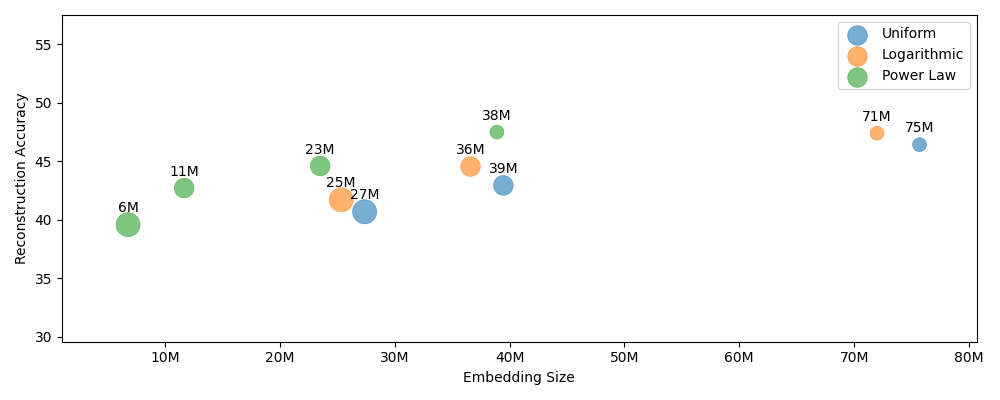}
\caption{Sensitivity analysis illustrating the trade-off between performance and model size for the embedding space partitioning mechanism. Each point represents a model trained with a different number of partitions (indicated by bubble size) and a distinct partitioning scheme (indicated by color).  } 
\label{fig_partstune}
\end{figure*}

\begin{table*}[t]
\caption{Instruction tuning results for tasks such as risk tagging, geographic span tagging, expense type tagging and identifying recurring transactions. Accuracy \% is used as the metric.}
\centering
\begin{tabular}{llrrrr}
\addlinespace[0.5ex]
\specialrule{0.8pt}{0pt}{0pt}
  \rule{0pt}{2.5ex} \textbf{Model Type} & \textbf{Model Size} &\textbf{Risk}  & \textbf{Geo} & \textbf{Expense} & \textbf{Recurrence} \rule[-1.2ex]{0pt}{0pt}  \\
\specialrule{0.8pt}{0pt}{0pt}
\addlinespace[0.5ex]
CatBoost & N/A & 60.9 & 71.6 & 50.0 & N/A \\
Multimodal Falcon Adapter Tuning   & 122M & \underline{95.1} & 86.7 & \textbf{70.3} & \underline{47.4} \\
Multimodal Phi Adapter Tuning   & 122M & \textbf{99.8} & \underline{88.6} & \underline{53.1} & \textbf{48.4} \\
Multimodal Falcon Full Tuning   & 1.3B & 82.9 & \textbf{92.4} & 50.0 & 47.3 \\
\addlinespace[0.5ex]
\specialrule{0.8pt}{0pt}{0pt}
\end{tabular}
\label{tab_intuneres}
\end{table*}

\subsection {Tabular Encoder Evaluation}
We compare different variants of the models to assess our solution. These include the \emph{Classical Embeds} setting, where vocabularies are treated traditionally; the \emph{Classical Loss} setting, which uses standard reconstruction loss instead of a composite loss; the \emph{Partitioned Embeds} setting, where accounts and embedding space are sub-divided; and the \emph{Partitioned + MetaCols} setting, which incorporates prelearned embeddings from account columns while excluding name columns during transaction representation learning. All models employ a bi-directional transformer with column masking in the standard BERT~\cite{devlin2018bert} configuration, except for the \emph{Autoregressive Mask} setting,  which uses a causal mask. The models are trained for $1$ epoch, with a learning rate of $0.0001$ using a cosine scheduler. For the partitioning embedder, we set $B=4$, $\alpha_v=-3$ and $\alpha_d=2.25$.  

The models are evaluated based on their ability to accurately reconstruct the masked columns. For instance, all \emph{Source} columns such as \emph{Source Account ID} and \emph{Source Country}, may be masked. Given the remaining columns, we assess whether the top three predictions of the masked columns match the true column values.

Table ~\ref{tab_pretrainres} presents the reconstruction accuracy for the different settings. The \emph{Partitioned + MetaCols} setting consistently ranks first or second across all columns. Although the \emph{Classical Embeds} setting performs well, it uses the highest number of parameters, making it impractical for real-world scenarios with millions of account identifiers. The partitioning mechanism, despite using $50\%$ fewer parameters, achieves better or comparable performance. Additionally, the improvement over the \emph{Classical Loss} setting indicates that incorporating metric learning in the loss function can be advantageous.

Figure ~\ref{fig_epochs} shows the results of an ablation study for different model sizes and training durations. There is significant improvement in reconstruction accuracy between the small (25M parameters) and medium (100M parameters) model sizes, but the gains plateau for the large model size (250M parameters with 18 layers and 1024 hidden size). Similarly, training for three epochs yields substantial gains over one epoch, while the improvements diminish beyond three epochs. Figure ~\ref{fig_partstune} illustrates the trade-offs involved in selecting different bin sizes ($B=2, 4, 6$), partitioning mechanisms (uniform, logarithmic, or power-law), and exponent values for the power-law. The power-law dynamics using a bin size of 4 seem to provide the desired balance between model size and performance.

\subsection {Multimodal Decoder Evaluation}
We employ a baseline CatBoost~\cite{dorogush2018catboost} classifier, and two different LLMs Falcon~\cite{almazrouei2023falcon} and Phi~\cite{li2023textbooks} to evaluate instruction tuning. The Recurrence task, which involves passing multiple transactions, is not suitable for non-sequential classifiers and is therefore excluded from the CatBoost evaluation for a fair comparison. For the full fine-tuning setting, we freeze the tabular encoder, while keeping the LLM unfrozen. Label prediction accuracy, consolidated for each task, serves as the evaluation metric. All LLMs were trained for $1$ epoch and used the $3$ epoch, $100M$ parameter, partitioned embeddings version of the tabular encoder.   

Table ~\ref{tab_intuneres} presents the results of instruction tuning. The proposed adapter tuning approach conveniently outperforms the other methods. Notably, despite using only a fraction of the parameters compared to the fully-tuned model, it achieves substantial gains. Figure ~\ref{fig_qual} provides qualitative examples of the predicted responses. The results are all well-formed, demonstrating the model's ability to distinguish between transaction and text modalities. The recurrence tagging task seems particularly hard, as it involves selecting row sentinels corresponding to the transaction modality.   

\subsection {Discussion}
The results from tabular encoder confirm that partitioning the embedding space in a non-uniform manner is an effective solution for handling large vocabularies of categorical items. Additionally, a multi-faceted loss function is essential for addressing representation localization issues. Despite the recent successes of autoregressive learning, bi-directional transformers remain relevant, especially when dealing with permutation invariant data such as tabular transactions. The multimodal decoder results indicate that parameter efficient tuning of the LLMs is the preferred solution for instruction tuning. However, challenges remain in handling complex tasks such as recurrence tagging, which requires deep cross-modal alignment.


\section{Conclusion}
Transaction data, although systematically organized with a well-defined structure, presents unique challenges distinct from conventional tabular formats. The presence of numerous sparsely distributed identifiers, a vast number of columns, and the necessity for precise numerical awareness demand scalable solutions tailored to these specific characteristics. In this paper, we addressed these challenges by developing embedders that reduce the complexity of large vocabularies, leverage inherent factorizations in tabular data, and remain robust to numerical variability. Additionally, we introduced an efficient mechanism to integrate transaction and text modalities using a language-based interface. While our solution was specifically designed for transaction data, it is also applicable to other similar formats.
In future work, we aim to delve deeper into the representation space and the internal behavior of the model to further enhance its capabilities and applicability.

\begin{ack}
This paper was prepared for information purposes by the Artificial Intelligence Research group of JPMorgan Chase \& Co and its affiliates (“JP Morgan”), and is not a product of the Research Department of JP Morgan.  J.P. Morgan makes no representation and warranty whatsoever and disclaims all liability for the completeness, accuracy or reliability of the information contained herein. This document is not intended as investment research or investment advice, or a recommendation, offer or solicitation for the purchase or sale of any security, financial instrument, financial product or service, or to be used in any way for evaluating the merits of participating in any transaction, and shall not constitute a solicitation under any jurisdiction or to any person, if such solicitation under such jurisdiction or to such person would be unlawful. © 2023 JP Morgan Chase \& Co. All rights reserved.
\end{ack}

\bibliographystyle{unsrt}  
\bibliography{references}

\end{document}